# Non-native speakers of English or ChatGPT: Who thinks better?

Mohammed Q. Shormani,
Department of English Studies, Ibb University, Ibb, Yemen
Department of English Studies, University of Cyprus, Nicosia, Cyprus
Email: shormani.mohammed@ucy.ac.cy
Orcid: https://orcid.org/0000-0002-0138-4793

**Abstract**

This study sets out to answer one major question: Who thinks better, non-native speakers of English or ChatGPT?, providing evidence from processing and interpreting center-embedding English constructions that human brain surpasses ChatGPT, and that ChatGPT cannot be regarded as a theory of language. Fifteen non-native speakers of English were recruited as participants of the study. A center-embedding English sentence was presented to both the study participants and ChatGPT. The study findings unveil that human brain is still far ahead of Large Language Models, specifically ChatGPT, even in the case of non-native speakers of an L2, here English. The study concludes that human brain's ability to process and interpret natural language data is unique and that ChatGPT still lags behind this human unique ability.

## 1. Introduction

Center-embedding as in (1) is a syntactic phenomenon in which a matrix clause contains several other relative (embedded) clauses. Put differently, center-embedding occurs when clauses are nested within other clauses, creating significant demands on working memory and sentence processing.

(1) a. The man that the soldier that the thief slapped deceived died.

    b. Men women children dogs bit like marry hate pets.

(1a) presents a triple center-embedding structure, and (1b) a quadrilateral one ((1b) is taken from Karlsson, 2007, p. 8, see also Frazier & Rayner, 1988). These constructions present a considerable difficulty for humans due to the complexity of structure they involve. These center-embedding structures, in fact, have been investigated since Chomsky and Miller (1963). They impose difficulty for human working memory, giving rise to processing difficulty, due perhaps to the memory load placed on the Faculty of Language (FL) while processing them. Additionally, (1b) is more difficult than (1a), and this difficulty lies in involving more embedded clauses, indefinite nouns/subjects, and absence of relative pronouns like *that*. Because of the absence of relative clauses, sentences like (1b) are said to be reduced relative clauses (Shormani, 2013).

Artificial intelligence (AI) has aroused much controversy among linguists and AI specialists over using AI models like ChatGPT and their capabilities. Much debate has been recently going on in this regard. On the one extreme, there are scholars who view AI models as incredibly able to do processing tasks like humans (see e.g., Piantadosi, 2023; Ambridge & Blything, 2024). For instance, Piantadosi (2023) claims that Large Language Models (LLMs) like ChatGPT are good language theories, and they can even "refute" Chomsky's generative approaches. On the other extreme, several scholars refute this claim and argue that, though these models do great tasks, AI models still fall short of reaching human brain State (Katz, 2012; Shormani, 2024a). Some scholars provide good evidence from natural language processing tasks that LLMs including ChatGPT cannot be considered language theories, because they still lag behind human brain State (see e.g., Zhong et al., 2023; Katzir, 2023).

The first idea behind the inception of AI is how to implement "human intelligence" in computers, making them think like humans. AI is defined as "making a machine behave in ways that would be called intelligent if a human were so behaving" (McCarthy et al., 1955, p. 11). It is a computer's underlying ability "to interpret external data correctly, to learn from such data, and to use those learnings to achieve specific goals and tasks through flexible adaptation" (Europe Business School, Paris see also Haenlein & Kaplan, 2019). The first definition ensues from scientific questions imposed in the 1950s, perhaps with Turing's (1950) stimulating question "Can machine think?". The second definition perhaps concerns the modern use of LLMs like ChatGPT in processing, computing and interpreting natural language data.

The term "artificial" in "artificial intelligence" implies that LLMs' "intelligence" is not real, and that they do not think like humans (see also Haenlein & Kaplan, 2019). However, there are scholars (Piantadosi, 2023; Ambridge & Blything, 2024) who see that these models think or process language data even better than humans. Thus, in this paper we aim to see to what extent this is true, examining ChatGPT's ability in processing center-embedding English sentences. We also aim to find out who is better in processing these constructions our participants, non-native speakers of English (NNSs), or ChatGPT. Our study recruited 15 NNSs; they are male and female. They are advanced learners of English as a second language. 14 of them are in the stage of writing their MA and PhD Theses, and 1 is a level-four student. We asked them one question involving a center-embedding structure. The same question was asked to ChatGPT.

Thus, the article is set up as follows. Section 2 briefly presents Language Faculty and center-embedding structures along with their salient syntactic properties. Section 3 spells out the current situation of LLMs and center-embedding, tackling studies in which AI specialists develop neural algorithms in LLMs to understand center-embedding structures and studies examining these models' abilities to understand, process, and interpret such structures. Section 4 outlines the methods followed in this study. Section 5 presents the results and discusses these results. Section 6 concludes the paper, providing some limitations and suggestions for future research.

**2. Language Faculty and center-embedding**

The generative enterprise has taken a biological and cognitive approach to the study of language, language faculty and language acquisition (Chomsky, 1995; Jenkins, 2000; Hauser et al., 2002; Shormani, 2016, 2017). For example, Hauser et al. (2002) discuss the unique features of human language, its evolution, proposing a framework for understanding language faculty, the responsible organ for language production and perception in terms of biological and cognitive mechanisms. They argued that there are two states of the Faculty of Language in the Broad Sense (FLB) and the Faculty of Language in the Narrow Sense (FLN), and that there are identified differences between them. FLB includes a combination of systems that support language: i) the sensory-motor system, which is responsible for speech and auditory processing, ii) the conceptual-intentional system, which is involved in meaning and intention, and iii) recursion, which allows the generation of infinite combinations of expressions from a finite set of elements. Put differently, FLN refers specifically to the core computational mechanism for recursion, which, they argue, is unique to humans. This mechanism enables the generation of hierarchical structures including embedding clauses within matrix clauses. It is also central to human linguistic capability. Hauser et al. (2002) emphasize the interdisciplinary nature of studying language evolution, involving fields like linguistics, evolutionary biology, and neuroscience. They focus on examining how FLB components evolved independently, exploring FLN's possible evolutionary origins and genetic basis. Between these two states, they suggest, LF sometimes fails to perform some linguistic tasks including comprehending complex structures, due mainly to the load placed on it, or its working memory. FL is in principle an intact organ, genetical wired in human genes, human and only human possess it.

Given the limited ability of FL and working memory in processing complex structures, humans encounter some difficulty in processing center-embedding structures, because their syntactically complex structure imposes difficulty for human working memory. Center-embedding has been investigated since Chomsky and Miller (1963), and developed in several works. For example, Frazier (1985) found that there is a breakdown of processing these structures. Dickey (1995) conducted a study in which reading time experiments were presented, which reveals that inserting an ungrammatical resumptive pronoun in the second of three noun gaps led to faster reading times. Additionally, Thomas (1995) investigates the cognitive processes involved in understanding sentences with center-embedding and self-embedding structures. These are types of recursive sentence formations where clauses are nested within one another, creating complex hierarchical structures. He found that center-embedding involves inserting subordinate clauses into a main clause. Self-embedding, on the other hand, occurs when multiple such embeddings of the same type are nested within each other, leading to more complex and often harder-to-process sentences. He also explores why deeply embedded sentences challenge human working memory and comprehension. Memory limitations and working memory capacity are key factors in making these structures difficult to process (see also Uehara & Bradley, 2002).

Karlsson (2007) studies the constraints on multiple center-embedding sentences, and their syntactic peculiarities. He examines the reasons why sentences with multiple center-embedded clauses like (1) are challenging to process for humans, despite being grammatically correct. In this study, Karlsson introduces the concept of *center-embedding ceiling*, where human cognition struggles to parse sentences with more than two levels of embedding. He further argues that the difficulty arises not merely from cognitive limitations but from structural constraints inherent to language, providing cross-linguistic examples. He argues that even though multiple center-embeddings are theoretically possible, they are rarely found in actual use due to both cognitive and communicative pressures. He concludes that center-embedding imposes limitations on working memory and principles of processing efficiency.

Additionally, Karlsson (2010) explores the limitations of human working memory when processing sentences involving complex syntactic structures, such as multiple center-embedded clauses. Karlsson perhaps builds on theories like the Syntactic Prediction Locality Theory (see e.g., Gibson, 1998), which suggests that increased distance between syntactic predictions and their resolutions leads to higher memory and integration costs. These costs become particularly problematic in cases of multiple center-embedding as in (1). Specifically, triple center-embedding structures demonstrate significant processing difficulty due to the high memory demands of maintaining unresolved syntactic predictions while simultaneously introducing new referents and dependencies. Karlsson argues that beyond a certain threshold, such structures exceed working memory capacity, rendering them effectively not processable in real-time. Karlsson's (2010) study concludes that working memory imposes a hard constraint on sentence complexity, highlighting the importance of syntactic simplicity and local dependencies in human language comprehension.

### 3. Center-embedding and LLMs

As has been stated above, center-embedding is a linguistic construction where relative clauses (full or reduced) are inserted into the middle of a sentence, creating a highly nested structure. In this section, we will focus on LLMs and their involvement in center-embedding phenomena. Center-embedding has been involved in natural language processing (NLP) in two aspects, resulting in ample studies. These studies can be classified into two types: i) studies concerning developing Neural Networking Algorithms (NNAs) with center-embedding algorithms, and ii) studies concerning testing LLMs' abilities to process and interpret these constructions. Concerning the first type, for instance, Jiang et al. (2024) have developed Prompt-based Method with Explicit One Word Limitation (PromptEOL), a method that leverages prompts for embedding sentences, and explores its efficacy both with and without fine-tuning.

PromptEOL is a technique that integrates in-context learning by providing specific prompts to LLMs, allowing them to generate sentence embeddings without additional parameter updates. When fine-tuning is applied, PromptEOL significantly enhances the performance of LLMs on various sentence-level tasks. However, without fine-tuning, PromptEOL surpasses state-of-the-art methods like SimCSE in semantic textual similarity benchmarks. The method benefits from model scaling, with embeddings improving as model size increases up to a point. However, there are a number of challenges: i) diminishing returns: the performance plateau for very large models raises questions about the scalability limits of LLMs in certain tasks, and ii) task-specific optimization: tailored approaches to optimize embeddings for different downstream applications were highlighted. They conclude that their proposal makes a strong case for the scalability and adaptability of LLMs in sentence embedding tasks. By introducing PromptEOL, it provides a pathway to harness the power of large models efficiently, bridging the gap between raw model size and practical performance.

Harris et al. (2024) explore a method to improve text embedding performance by preprocessing input text using LLMs, specifically ChatGPT 3.5. They consider that this approach is crucial for various NLP tasks, but there are a number of factors that limit its functionality such as vocabulary, lack of context, and grammatical errors. The approach involves enriching text with context, correcting grammatical errors, disambiguating terms, and including relevant metadata before generating embeddings. The aim is to enhance the performance of embedding models in downstream tasks such as classification and clustering. The study evaluates the method on three datasets: Banking 77 Classification, TwitterSemEval 2015, and Amazon Counter-Factual Classification. It uses metrics like cosine similarity and accuracy. Results show notable performance improvements, especially on the TwitterSemEval dataset, where the proposed technique achieved a significant leap from the previous best performance (85.34 vs. 81.52 on the Massive Text Embedding Benchmark). However, improvements on other datasets were mixed, underscoring the dependency on dataset characteristics.

The second type of studies is concerned with examining LLMs to process and interpret center-embedding structures. For example, Kodner et al. (2024) provide a reply to Piantadosi's assertion that modern LLMs challenge Chomsky's linguistic theories. The authors defend the relevance of generative linguistics and argue for its continued importance in understanding human language. They focus on four issues: i) the data gap: they emphasize the disparity between the vast data requirements of LLMs and the minimal exposure young children need to acquire language. This highlights the unique mystery of human language acquisition that generative linguistics seeks to explain, ii) artificial vs. natural insights: they draw an analogy between LLMs and airplanes, suggesting that while airplanes reveal much about engineering, they offer little insight into natural avian flight. Similarly, LLMs' functioning may not elucidate the cognitive mechanisms underlying human language, iii) limits of LLMs as scientific theories: in this regard, the authors argue that scientific theories require interpretable explanations, not just predictive accuracy. Since LLMs lack explicit theoretical frameworks, they cannot replace linguistic theories, and iv) the necessity of independent linguistic theories: evaluating LLMs' capabilities still depend on understanding human linguistic capacities. Generative linguistics provides a robust framework for such evaluation, underscoring its indispensable role in linguistic science. Kodner et al. conclude that generative linguistics will remain crucial in advancing our understanding of language, despite technological progress in computational models.

Dentella et al. (2024) assert that language is not an attribute that can be ascribed to LLMs. In their experiment, seven LLMs fail to respond to simple questions based on textual input including examples such as (2).

(2) a. "John deceived Mary and Lucy was deceived by Mary"

  b. "In this context, did Mary deceive Lucy?"

While human participants succeeded in this task, though they sometimes err, not only did LLMs fail, but they also kept doing the same errors, which is not human. If a human makes a mistake one time, s/he does not repeat it if his/her attention is paid to that error. They recruited 400 English native speakers and GPT-3 and GPT-3.5. They concluded that LLMs lack a compositional operator that integrates and regulates grammatical and semantic information.

Katzir (2023) criticizes the claim that LLMs, such as GPT-3 and similar models, serve as robust theories of human linguistic cognition. In fact, Katzir argues against Piantadosi's (2023) argument that LLMs outperform generative linguistics in explaining human language cognition. He provides objections in relation to three phenomena: i) competence vs. performance: Katzir emphasizes that LLMs lack the ability to distinguish between linguistic competence and performance. The former refers to the native speaker's underlying knowledge of his/her language, and the latter refers to his/her use of language in actual situations (Chomsky, 1965). In Chomsky's words, there is "a fundamental distinction between competence (the speaker-hearer's knowledge of his language) and performance (the actual use of language in concrete situations)" (p. 4). In this sense, there is perhaps a distinction between human and machine or computer, in that competence is a human attribute, specifically a characteristic of the human brain whose "linguistic performance" is an indication of the underlying linguistic competence. On the other hand, computers or LLMs, in specific terms, can have performance, i.e. producing a similar phrase/sentence or even a text based on the data (i.e. the corpus) they have been trained on, which does not entail that they have "competence" like humans (see also Kaufer, 1979). If this is on the right track, it then follows that the distinction between competence and performance is central to understanding human linguistic behavior, as humans often struggle with sentences due to processing limitations, not a lack of competence. In contrast, LLMs' errors reflect deficits in their statistical learning mechanisms, not resource constraints, ii) likelihood vs. grammaticality: Katzir considers that humans can discern grammatical but unlikely sentences from likely but ungrammatical ones. This is perhaps due to the neurological mechanism the brain working mechanism is based on. In terms of connectionist models of the FL models, the human brain tries to identify the functional task through which it comprehends or produces speech (Arbib & Caplan 1979; Nelson, 1979; Arbib, 1982), and iii) typological universals: LLMs do not inherently explain cross-linguistic typological universals, which generative linguistics seeks to address LLMs may be insufficiently biased towards these universals, making them implausible models for understanding human linguistic diversity.

Additionally, Zhong et al. (2023) evaluates ChatGPT's natural language understanding (NLU) capabilities against fine-tuned BERT models (BERT-base, BERT-large, RoBERTa-base, and RoBERTa-large) using the GLUE benchmark. They found that ChatGPT's performance is comparable to BERT-base (78.7% vs. 79.2% average score) but lags behind more advanced models like RoBERTa-large (87.8%) (see also Ettinger, 2020). However, ChatGPT struggles with paraphrase detection and semantic similarity underperforming BERT-base by as much as 24% in some cases. The study suggests that while ChatGPT is versatile, it still lags behind highly specialized fine-tuned models in specific NLU tasks. This underscores the complementary nature of task-specific fine-tuning and the broad generalization abilities of LLMs including ChatGPT. This aspect can also be contrasted with human brain, a species-specific property, having a unique working mechanism (see also Chomsky, 2009; Berwick & Chomsky, 2016).

Bearing the above review in mind, the present study seeks to answer the following questions:

1. Who thinks better, NNSs or ChatGPT?
2. To what extent do NNSs understand, interpret, and perceive center-embedding structures?
3. To what extent do LLMS, specifically ChatGPT understand, interpret, and perceive center-embedding structures?

**4. Methods**

**4.1. The study data**

The study data consist of a sentence involving center embedding phenomenon, namely *The man that the soldier that the thief slapped deceived died*. The same sentence was used as an instrument for judgement, i.e. we asked ChatGPT to judge whether the sentence is grammatical. We used only one center-embedding sentence as the data of the study for two reasons: i) if human participants err in one such sentence, they will commit the same errors in similar sentences, and the same thing can be said also about ChatGPT, and ii) to avoid redundancy ensuing from discussing and/or interpreting the same data that contain the same errors either by human participants or ChatGPT.

**4.2. Participants**

This study involves 15 participants, they are male and female. They are NNSs- they are advanced L2 English students. Two are PhD students, doing their PhD Dissertations, and twelve are MA students, doing their MA Theses, and one is a level-four student.

**4.3. Procedure**

The English sentence was sent to the participants via WhatsApp. The question was:

"What each did in the following sentence:

"The man that the soldier that the thief slapped deceived died."

The man:
The soldier:
The thief:

I am doing a research and I need your answer."

The participants were given 24 hours to answer the question. The time they submitted their answers is somehow different for each one, some answered immediately, some took about two hours, and two submitted their answers next day, but within the time allotted. Those who took a long time had not seen our message. We did not respond to some of them asking for further contextual clues. Our response to them was "We cannot give you any further clarification, because doing so will affect the study results." To refer to the participants, we did not use their real names, instead we used numbers (1-15). The same question was asked to ChatGPT. We utilized ChatGPT-3.5, because it was the only available free version for us while conducting the study. ChatGPT's answer was recorded on October 3, 2024. Figures 1 and 2 display ChatGPT's answers for both the processing and judgement tasks:

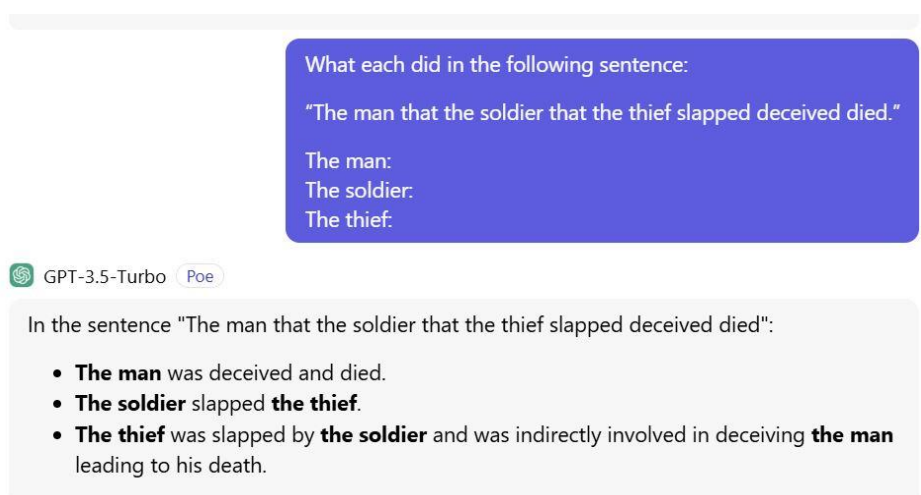

**Figure 1: ChatGPT's processing task**

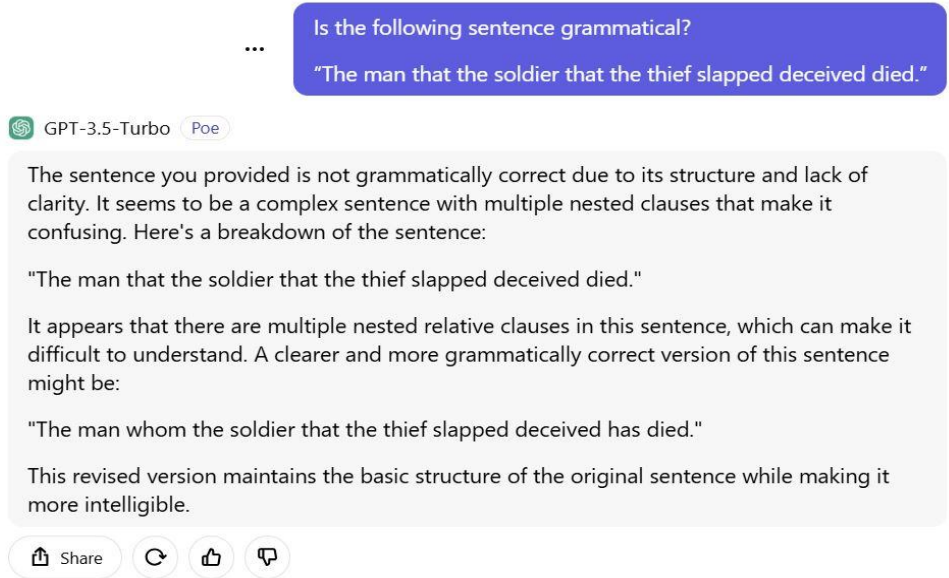

**Figure 2: ChatGPT's judgment task**

**5. Results and discussion**

**5.1. Results**

In this section, we will tabulate the NNSs' and ChatGPT's responses. Table 1 showcases these responses. It displays the answers of 15 participants for the three entities meant, viz., *the man, the soldier* and *the thief*. It also presents ChatGPT's responses for these entities.

**Table 1: NNSs' and ChatGPT's responses (for *the man, the soldier and the thief*)**

| Participant | The man | The soldier | The thief |
|---|---|---|---|
| 1 | died | deceived | slapped |
| 2 | died | slapped | slapped |
| 3 | died | deceived | slapped |
| 4 | died | deceived | slapped |
| 5 | died | deceived | slapped |
| 6 | died | deceived | slapped |
| 7 | slapped | deceived | died |
| 8 | nil | nil | nil |
| 9 | died | slapped | deceived |
| 10 | died | deceived | slapped |
| 11 | died | deceived | slapped & deceived |
| 12 | nil | nil | nil |
| 13 | died | deceived | slapped |
| 14 | died | deceived | slapped |
| 15 | died | deceived | slapped |
| ChatGPT | died | slapped | deceived |

**5.2. Discussion**

As Table 1 showcases, almost all the participants made good judgements on the stimulus sentence, and stated what *the man, the soldier,* and *the thief* each did, while ChatGPT failed to do so. We will just present those who got it wrong. Participant 2 has 2 mistakes, viz. s/he didn't get the correct` answer for the *soldier* and *thief*. Participant 7 got it right for the soldier, but s/he mixed between the *man* and *thief*. Participant 9 did it right for the *man*, but s/he mixed between what the *solider* and *thief* each did. Participants 8 and 12 left it undone. But we consider it as if they did it wrong. So, we have 12 incorrect answers out of the 45 total answers for the 3 entities, i.e. the *man, soldier* and *thief*. In human answers, we have 26.7% incorrect answers and 73.3% correct answers. 5 participants (including those who left it undone) have made mistakes, i.e. 33.3%, 10 participants, viz. 66.7% provided correct answers. As for ChatGPT, it only provided one correct answer, i.e. only for the *man*.

The way ChatGPT answers the question is strange, because it adds information messing up things. Human participants also add more information, say, for instance, stating the object who receives the action, but they do not mess up things. Put simply, although the question was clear that what is needed is only what each subject did, no matter what action the object received, humans added reasonable information while ChatGPT provided nonsensical information. That is, ChatGPT errs even in this (extra) information. The following is ChatGPT's exact answer:

> The man was deceived and died.
> The soldier slapped the thief.
> The thief was slapped by the soldier and was indirectly involved in deceiving the man leading to his death.

In ChatGPT's answer, we find that it does not understand center-embedding structures though it has been developed for that as we have seen so far. Put differently, ChatGPT has incorporated algorithms to process center-embedding sentences (see e.g., Jiang et al., 2024; Harris et al., 2024; and references therein), but it fails to process and interpret our sentence. In *The man was deceived and died,* we know that *the man* died, but he deceived no one. Additionally, ChatGPT's answer with regard to the *soldier*, viz., *The soldier slapped the thief* is completely incorrect. The third part of its answer is just nonsense.

ChatGPT's failure to answer the question urges us to examine its grammaticality judgement. Put simply, if ChatGPT's answer was as such, one may wonder what and how its grammaticality judgement is. We asked it the following question to find out its judgment in terms of grammaticality.

> User: Is the following sentence grammatical?
>
> "The man that the soldier that the thief slapped deceived died."

> The sentence you provided is not grammatically correct due to its structure and lack of clarity. It seems to be a complex sentence with multiple nested clauses that make it confusing. Here's a breakdown of the sentence:
>
> "The man that the soldier that the thief slapped deceived died."
>
> It appears that there are multiple nested relative clauses in this sentence, which can make it difficult to understand. A clearer and more grammatically correct version of this sentence might be:
>
> "The man whom the soldier that the thief slapped deceived has died."
>
> This revised version maintains the basic structure of the original sentence while making it more intelligible.

This finding makes our study partly in contrast with Katzir (2024): while ChatGPT judges Katzir's sentence as "grammatically correct", in our study it judges the sentence as "not grammatically correct". This also provides another piece of evidence that ChatGPT is inconsistent. However, all NNSs consider it grammatical. Although we have not asked them directly about it., we, in fact, infer their grammaticality judgement indirectly, because none of them said the sentence is "wrong/ungrammatical". As for ChatGPT's answer in relation to misunderstanding center-embedding structures, our study is in line with Katzir's (2023) findings. ChatGPT in both studies lags behind expected performance. Both studies demarcate its poor performance, which needs further specialized development. Our study also supports Dentella et al.'s (2024) study in which human participants outperformed ChatGPT. The difference between our study and theirs concerns the type of participants - while our participants are NNSs, their participants are native speakers of English.

The fact that LLMs like ChatGPT fail to process center-embedding structures could be accounted for if we know the basis on which they function. They largely depend on statistics and statistical procedures. Put differently, LLMs working mechanism is based on probabilities, i.e. they just "guess" or "predict" the n-gram word, which is not always error-free. This deficiency in mechanism could be attributed to their "competence", viz., their underlying ability to point out/choose the correct word, not predict or guess it. LLMs may also lack "competence" in the linguistic technical sense. In contrast, native speakers of an L may fail to do a linguistic task, as in the case of Dentella et al. (2024), and their failure is not ascribed to "deficiency" in their competence, but to performance, which is particularly ascribed to psychological factors such as slips of tongue, fatigue, or not paying attention, which are all nonlinguistic (see also Chomsky, 2009). Likewise, if we assume that NNSs have built a "perfect" linguistic system, i.e. they mastered the L2 linguistic system, and if they fail to do a linguistic task, this failure cannot be ascribed to a "deficiency" in their linguistic competence, but rather to their performance, i.e. their ability to use or judge a piece of language.

These facts have been noted long back in generative enterprise. For instance, Chomsky (1975) points out that generative approach to the study of language "contrasts with a statistical approach that leads to an ordering of sequences from more to less probable, rather than a sharp division into two classes within which no such gradations are marked". His nonsense famous phrase is a good case in point here. The sentence *Colorless green ideas sleep furiously* and its opposite sequence *furiously sleep ideas green colorless* as described by him "are not distinguished by their assigned probabilities. If probability is to be based on an estimate of frequency in some English corpus, then this probability will be so in both cases" (p. 145). Additionally, corpus-based probabilities and their low order lack explanatory adequacy due to their being "predictional" like when, for instance, a sentence appears in a linguistic corpus (Kaufer, 1979). We can conclude that our analysis supports studies that critique LLMs, in that LLMs, though powerful tools, fail to account for the dynamic interplay between competence and performance that is central to human linguistic behavior (see also Katzir, 2023).

**6. Conclusions and limitations**

To conclude, we have examined the ability of both NNSs and ChatGPT in processing and interpreting center-embedding English sentences. We have recruited 15 NNSs as participants, advanced L2 English students, BA, MA, and PhD students. They outperformed ChatGPT in both processing/interpreting and judging the grammaticality of the given center-embedding sentence. Our study, thus, highlights the fact that human brain of (advanced) NNSs, like that of native speakers of English (as in the case of Dentella et al., 2024), process and interpret a complex English structure, viz., a center-embedding sentence far better than LLMs like ChatGPT, and that these LLMs cannot be considered powerful theories of language, as recently claimed by some researchers (Piantadosi, 2023; Ambridge & Blything, 2024). Another conclusion that can be drawn here concerns NNSs and their linguistic competence. Given the percentage of their answers to the center-embedding sentence presented to them, viz., 73.3%, we can conclude that their linguistic system of English is almost like that of native speakers. Those who erred (including those who did not provide answers) if we assume that NNSs have built a "perfect" linguistic system, i.e. they mastered the L2 linguistic system, and if they fail to do a linguistic task, this failure cannot be ascribed to a "deficiency" in their linguistic competence, but rather to their performance, i.e. their ability to use or judge a piece of language.

The study findings unveil that NNSs performed better than ChatGPT, which indicates that: i) LLMs need further developments, and ii) human brain (even in the case of NNSs) surpasses LLMs. Although center-embedding imposes difficulty for native speakers' FL working memory (see e.g., Dickey 1995), the fact that our participants, who are NNSs, performed better than ChatGPT has several implications for both generative linguistics enterprise and AI technology, ChatGPT, in particular, the most prominent of which are: i) generative linguistics enterprise: NNSs' competence mirrors that of native speakers. This in turn indicates that once a learner develops and masters a linguistic system of an L2, here English, this linguistic system does not differ much from that of the native speaker of this L2 (see e.g., Cook, 1983; White 2003; Shormani 2014a & b, among others). This is also further evidence of Chomsky's conceptions of Language Faculty, Universal Grammar, Genetic Endowment, and Language Innateness (Chomsky, 1957, 2001, 2008, 2021, see also Shormani, 2016, 2023, 2024b). And ii) AI technology: AI LLMs still need further developments to overcome these challenges. The study also refutes Piantadosi's (2023) arguments that LLMs are good theories of language, and that these models cannot refute Chomsky's generative enterprise. It also supports Dentella et al.'s (2024) findings that language is an attribute of humans and only humans, a species-specific property, and that it cannot be attributed to LLMs.

The study findings also provide potential insights into LLMs' working mechanism. ChatGPT in our study errs in both the processing task and the judgement task. However, this seems not strange given the fact that LLMs base their predictions purely on statistical likelihood, often favoring plausible continuations that may be grammatically incorrect, which demonstrates a fundamental limitation compared to human cognition (see also Katzir, 2023). It could be argued that while LLMs may excel as engineering tools, their design and functionality fall short of providing a scientific model for human linguistic cognition. Our aim in the paper has been to pinpoint the actual State of Human brain (even in NNSs) and that of LLMs. Our stimulus is a center-embedding sentence. This sentence is used to highlight fundamental differences between human linguistic cognition and LLMs' capabilities, which are basically statistical. The fact that humans often struggle with center-embedded sentences like (1) above, cannot be ascribed to a deficiency in their genetic linguistic knowledge, but it is simply due to the fact that these structures tax working memory, and the more the center-embedding sentence gets complex, the more the load is placed on FL. Therefore, humans' failures are attributed to performance limitations rather than a lack of understanding of syntactic rules (or competence). However, LLMs' either successes or failures are based on their statistical training and inherent model structure. Their competence directly reflects their behavior - errors are not due to transient resource constraints but to the limitations of their learned representations (Katzir, 2023). According to Katzir, while humans may initially struggle with center-embedded sentences, they can often parse them correctly with additional time or contextual clues. This adaptability is a hallmark of human cognition and is absent in LLMs. LLMs lack mechanisms for resource-based recovery. LLMs' ability to distinguish between competence and performance limits their usefulness as models for human cognition, hence cannot be regarded as powerful language theories. These findings are in line with Katzir's (2023) study. Like Katzir, we utilized center-embedding to examine whether LLMs can capture the nuances of how humans process, interpret, and judge complex linguistic structures.

This study, however, has some limitations. The first limitation concerns the type of sentences involved. A comprehensive study might involve other types of complex English sentences such as those involving DP-islands, multiple wh-questions, anaphora and weak/strong crossovers to assess both NNSs' and ChatGPT's capabilities in a wider context. The second limitation that can be tackled here has to do with the version of ChatGPT, viz., -3-5. A broader study could utilize ChatGPT-4. The latter is said to be more developed in functionality and features, and utilizing it could widen the study scope and purpose.

**Acknowledgements**

The author would like to very much thank the participants of the study for their time and valuable contribution.

**Ethical statement**

All the participants provided their consent for publishing and replicating their data.

**Data availability**

The data underlying the results of this study are available from the author upon request.